\documentclass[letterpaper, 10 pt, conference]{ieeeconf}  

\IEEEoverridecommandlockouts                              

\usepackage{graphics} 
\usepackage{epsfig} 
\usepackage{mathptmx} 
\usepackage{times} 
\usepackage{amsmath} 
\usepackage{amssymb}  
\usepackage{subfigure}
\usepackage{parskip}
\usepackage{multirow} 
\usepackage{booktabs}
\usepackage{siunitx}
\usepackage{makecell}
\usepackage{caption}
\title{\LARGE \bf
STDG: Semi-Teacher-Student Training Paradigram for Depth-guided
One-stage Scene Graph Generation
}

\author{Xukun Zhou$^{1}$, Zhenbo Song$^{2}$, Jun He$^{1}$, Hongyan Liu$^{3}$ and Zhaoxin Fan$^{1,4}$
\thanks{$^{1}$Renmin University of China
        {\tt\small \{xukun\_zhou,hejun, fanzhaoxin\}@ruc.edu.cn}}%
\thanks{$^{2}$Nanjing University of Science and Technology, China
        {\tt\small songzb@njust.edu.cn}}%
\thanks{$^{3}$Tsinghua University, China
{\tt\small hyliu@tsinghua.edu.cn}}
\thanks{$^{4}$Psyche AI Inc, China
{\tt\small fanzhaoxin@psyai.net}}
}

\begin{document}

\maketitle
\thispagestyle{empty}
\pagestyle{empty}

\begin{abstract}


Scene Graph Generation is a critical enabler of environmental comprehension for autonomous robotic systems. Most of existing methods, however, are often thwarted by the intricate dynamics of background complexity, which limits their ability to fully decode the inherent topological information of the environment. Additionally, the wealth of contextual information encapsulated within depth cues is often left untapped, rendering existing approaches less effective. To address these shortcomings, we present STDG, an avant-garde Depth-Guided One-Stage Scene Graph Generation methodology. The innovative architecture of STDG is a triad of custom-built modules: The Depth Guided HHA Representation Generation Module, the Depth Guided Semi-Teaching Network Learning Module, and the Depth Guided Scene Graph Generation Module. This trifecta of modules synergistically harnesses depth information, covering all aspects from depth signal generation and depth feature utilization, to the final scene graph prediction. Importantly, this is achieved without imposing additional computational burden during the inference phase. Experimental results confirm that our method significantly enhances the performance of one-stage scene graph generation baselines.

\end{abstract}

\section{Introduction}
Scene graph generation, a task first introduced by Johnson et al. \cite{johnson2015image}, transforms an input image into a structured scene graph. This graph encapsulates the objects, attributes, and relationships within the scene, offering a detailed representation of the environment. The utility of scene graph generation spans across multiple domains. It bolsters scene understanding \cite{amodeo2022og}, aids in visual question answering \cite{wu2017visual,qian2022scene,hildebrandt2020scene}, and supports advancements in autonomous driving \cite{yu2021scene,malawade2022spatiotemporal}. The application of scene graphs is even particularly significant in the domain of robotics. In particular, in robotic navigation, the scene graphs enhance decision-making \cite{amiri2022reasoning}, facilitate the execution of complex tasks \cite{zhu2022scene}, and enable interaction with environmental objects \cite{FAN2022102304}. Hence, scene graph generation is a crucial tool in the enhancement of robotic capabilities.

Existing scene graph generation methodologies predominantly fall into two categories: two-stage approaches and one-stage approaches. The two-stage approach first employs an object detection network to identify and locate objects within an image \cite{yang2022panoptic,Li_2022_CVPR,Lin_2022_CVPR}. Once objects are identified, the information is used as ground truth input for a separate relationship prediction network \cite{dong2022stacked,li2022devil}. This network predicts the relationships between the various objects. Despite its effectiveness, this approach requires training two separate networks, leading to significant time and computational costs. In contrast, the one-stage approach consolidates the object detection and relationship prediction tasks into a single network. Newell et al. \cite{newell2017pixels} first propose this concept, which is later expanded by Liu et al. \cite{liu2021fully} into the Fully Convolutional Scene Graph Generation (FCSGG) method. This method models relationships as integral distribution maps on the image. Adaimi et al. \cite{adaimi2023composite} further advance this approach by optimizing the matching relationships between different objects, thereby enabling the generation of more accurate scene graphs while maintaining computational efficiency.
  
Despite advancements in one-stage scene graph generation, these methods continue to face substantial challenges. Two key difficulties lie in the ability to filter out complex backgrounds from RGB images and the need for a deeper understanding of scene topology. The rich color information in RGB images often complicates accurate modeling of object relationships, particularly for simpler models. One intuitive solution that emerges to address these challenges is the use of depth information. Depth data, unaffected by the influence of color, provides a potential pathway to more accurate scene graph generation. Not only does depth data provide an alternative to the color complexities of RGB images, but it also delivers rich 3D information. This 3D information could significantly enhance the network's understanding of the scene's topology, providing a more direct approach to modeling objects, attributes, and relationships. Moreover, depth information could potentially allow the network to bypass complications such as occlusion, which are common when predicting relationships within 2D images. By considering the depth of different objects, the network could gain a better understanding of which objects are in front of others, improving the accuracy of its predictions. This insight into the potential use of depth information prompts the question: Can depth information be leveraged to guide one-stage scene graph generation?



   Motivated by this, we propose a novel Depth-Guided One-Stage Scene Graph Generation method, termed as STDG. The innovation of STDG lies in the use of predicted depth as a robust guiding signal to enhance scene graph generation performance, with negligible additional computational cost. In STDG, we first introduce a Depth Guided HHA Representation Generation Module. In this module, we use an off-the-shelf depth predictor to estimate depth information from monocular images, which is then transformed into a novel HHA structure. This HHA structure is capable of better reflecting the topology of the scene and facilitates easier feature learning by the network. Subsequently, we train a Depth Guided Semi-Teaching Network Learning Module. Within this module, we first train a teacher network for scene graph generation using the HHA. We then utilize this teacher network to "teach" a student network on how to learn to predict scene graphs from images (as opposed to from HHAs). During this "teaching" process, we only "inherit" key aspects of the teacher that are most beneficial for scene graph generation, making this a semi-teaching network. Finally, we adopt a novel Depth Guided Scene Graph Generation Module as prediction heads to simultaneously predict all information of the scene graph.   Upon completion of the training phase, only the student network is required for inference, eliminating the need for further depth prediction. This results in a fast and lightweight inference solution that enhances operational efficiency.

To demonstrate the effectiveness of STDG, we conduct experiments on the well-known VG-100k dataset \cite{dosovitskiy2021image}. Experimental results demonstrate that our method significantly improves the performance of one-stage scene graph generation.


   
\begin{figure*} 
\centering 
\includegraphics[width=1.0\textwidth]{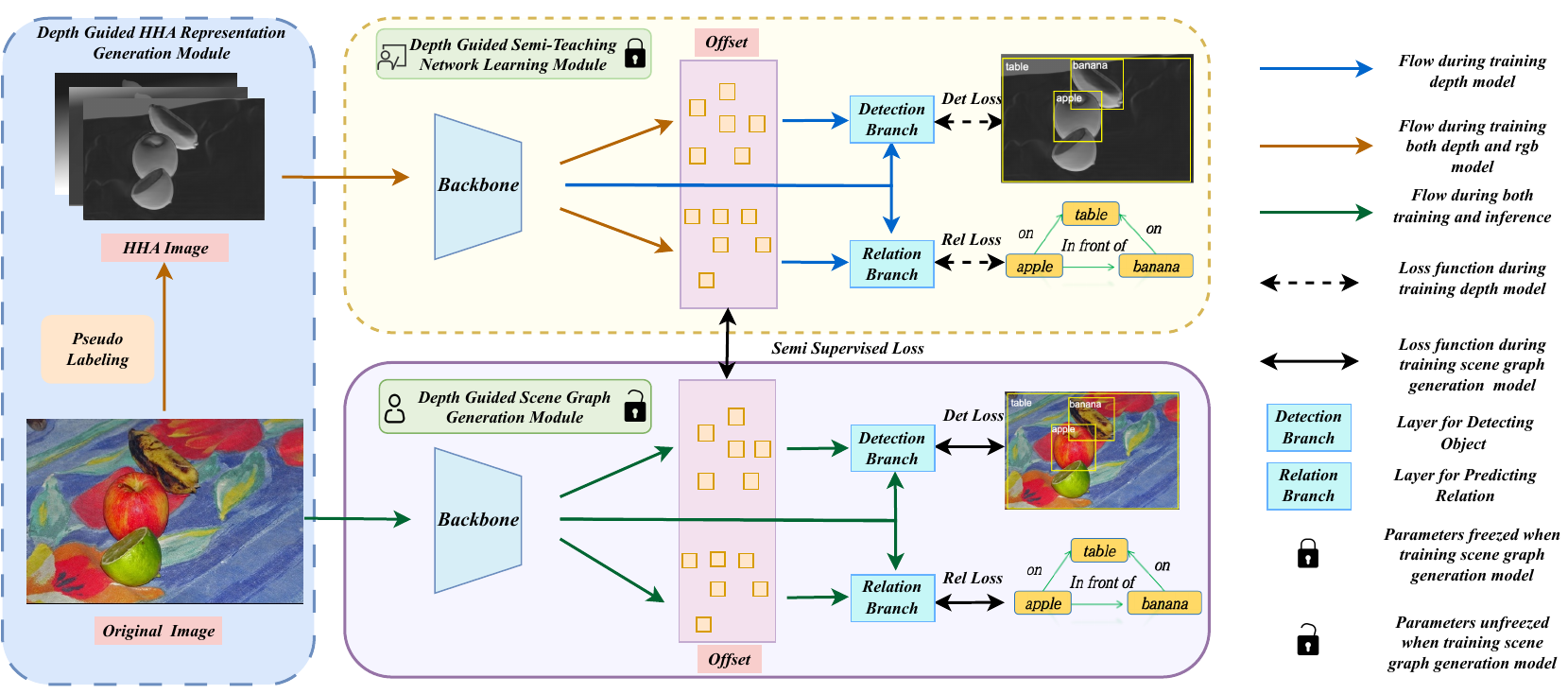} 
\caption{Pipeline of the STDG method. The process commences with the \textit{Depth Guided HHA Representation Generation Module}, creating an efficient HHA structure to facilitate the learning process. Subsequently, the \textit{Depth Guided Semi-Teaching Network Learning Module} is trained, which later instructs the \textit{Depth Guided Scene Graph Generation Module} on object prioritization. The terms \textit{Det Loss} and \textit{Rel Loss} denote the object detection and relationship losses in the scene graph generation process, respectively.} 
\label{Fig.Model} 
\vspace{-0.25in}
\end{figure*}

\section{Related Work}
\subsection{Two-stage Scene Graph Generation}
Scene graph generation traditionally adopts a two-stage approach. First, an object detection model, such as the Feature Pyramid Networks (FPN) \cite{Lin_2017_CVPR} or DEtection TRansformer (DETR) \cite{carion2020end}, is trained to identify the positions and categories of objects in an image. After detecting the objects, a complete graph is assumed to exist, representing their relationships. This graph is then processed by a relation classification network to determine the type of each relationship \cite{li2017scene,yang2018graph,tang2020unbiased,Gu_2019_CVPR}.Efforts to refine this approach can be categorized into two main areas. First, several works have proposed methods for enhancing the extraction of visual information. For instance, the use of global information to strengthen object representation has been suggested \cite{cong2021spatial, ji2020action, teng2021target, lu2021multi}. Furthermore, there has been advocacy for improving the convolutional network's focus on region sizes to propagate information between different objects \cite{chen2019knowledge, ji2020action, cong2021spatial}. The second area of focus has been the mitigation of the long-tail effect during the prediction process. This involves addressing overfitting to the dataset distribution during training. Tang et al. \cite{tang2020unbiased} suggested enhancing the relationship representation by adjusting the average relationship probability distribution, while Chiou et al. \cite{chiou2021recovering} proposed dynamically adjusting the loss for different relationships during training to enhance the learning of rare samples. Although traditional two-stage approaches to scene graph generation have shown promising results, they typically suffer from slower processing speeds and the potential accumulation of errors due to the necessity of two separate network stages. To circumvent these shortcomings, this paper explores the more efficient one-stage method.

\subsection{One-stage Scene Graph Generation} 
As previously discussed, two-stage methods for scene graph generation are burdened by slower processing speeds and the potential for error accumulation. To address these issues, one-stage methods have been introduced. One-stage methods primarily focus on relationship modeling without accounting for an object's position. Pioneering work by Newell et al. \cite{newell2017pixels} employed dual heatmaps to independently predict the positions of objects and relationships. This was followed by feature extraction corresponding to the identified relationship location, and computation of their interactions. FCSGG \cite{liu2021fully} expanded on this concept, initially considering human keypoints matching for scene graph object handling. Subsequent methodologies, such as CoRF\cite{adaimi2023composite}, further developed this idea, utilizing a random field as the relationship classifier. Despite these advancements, one-stage methods remain in their infancy and are still under active exploration. Furthermore, we observe that current one-stage methods are heavily influenced by complex color backgrounds, failing to adequately perceive scene topology and 3D layout. To address these challenges, this paper proposes a novel depth-guided one-stage scene graph generation method.

\subsection{Depth Estimator and Utlization} 
Predominantly, there are two methods to estimate depth information in the real world: stereo depth estimation and monocular depth estimation. Stereo depth estimation, such as the method proposed by Badki et al.\cite{badki2020bi3d}, utilizes a dual-camera setup to capture a single scene, computing depth by comparing the disparity between the two captured images. On the other hand, monocular depth estimation, like the approach introduced by Zhao et al.\cite{zhao2020monocular}, leverages a single camera and machine learning techniques to infer depth information from visual cues in a single image. Depth information has found extensive application across various domains, significantly enhancing performance. For instance, Gautier et al.\cite{gautier2011depth} used depth for improved object detection, while Ding et al.\cite{ding2020learning} utilized depth information for better scene understanding in autonomous driving.  Inspired by these developments, we propose to integrate depth information into scene graph generation to filter out unnecessary background information in a depth guided semi-teaching way. Given that scene graph generation tasks provide only a single image for each scene, we chose to employ a monocular depth estimator to generate pseudo labels, following the methodology of monocular depth estimation studies. To the best of our knowledge, we are the first to propose a depth-guided scene graph generation method.



\section{Method}
   
\subsection{Overview}
    The task of one-stage scene graph generation is defined as follows. Given an image $I$, a network $G$ is used to generate detections for $n$ objects and relationships between objects within $m$ relationship categories. Mathematically, we express this as:

\begin{equation}
G(I) = (Det^n, Rel^m)
\end{equation}

where $Det^n$ represents the detection results of $n$ objects and $Rel^m$ denotes the relationships between objects, categorized into $m$ kinds of relationships. 

Fig.\ref{Fig.Model} illustrates the pipeline of our proposed method, termed STDG. At the heart of our approach is a depth-guided method. Firstly, we introduce a Depth Guided HHA Representation Generation Module. In this module, an off-the-shelf depth predictor is used to estimate depth information from monocular images, which is transformed into a novel HHA structure. This HHA structure better reflects the scene's topology and facilitates easier feature learning by the network. Next, we train a Depth Guided Semi-Teaching Network Learning Module. Within this module, we initially train a teacher network for scene graph generation using the HHA. This teacher network is then utilized to instruct a student network on how to learn to predict scene graphs from images, as opposed to from HHAs. During this teaching process, we only inherit key aspects of the teacher that are most beneficial for scene graph generation, creating a semi-teaching network. Finally, we adopt a novel Depth Guided Scene Graph Generation Module as prediction heads to simultaneously predict all information of the scene graph. Upon completion of the training phase, only the student network is used for inference, eliminating the need for further depth prediction. This results in a fast and lightweight inference solution that enhances operational efficiency. In the following sections, we provide detailed descriptions of the three modules we propose: the Depth Guided HHA Representation Generation Module, the Depth Guided Semi-Teaching Network Learning Module, and the Depth Guided Scene Graph Generation Module.

\begin{figure}
    \centering
    \includegraphics[width=.45\textwidth]{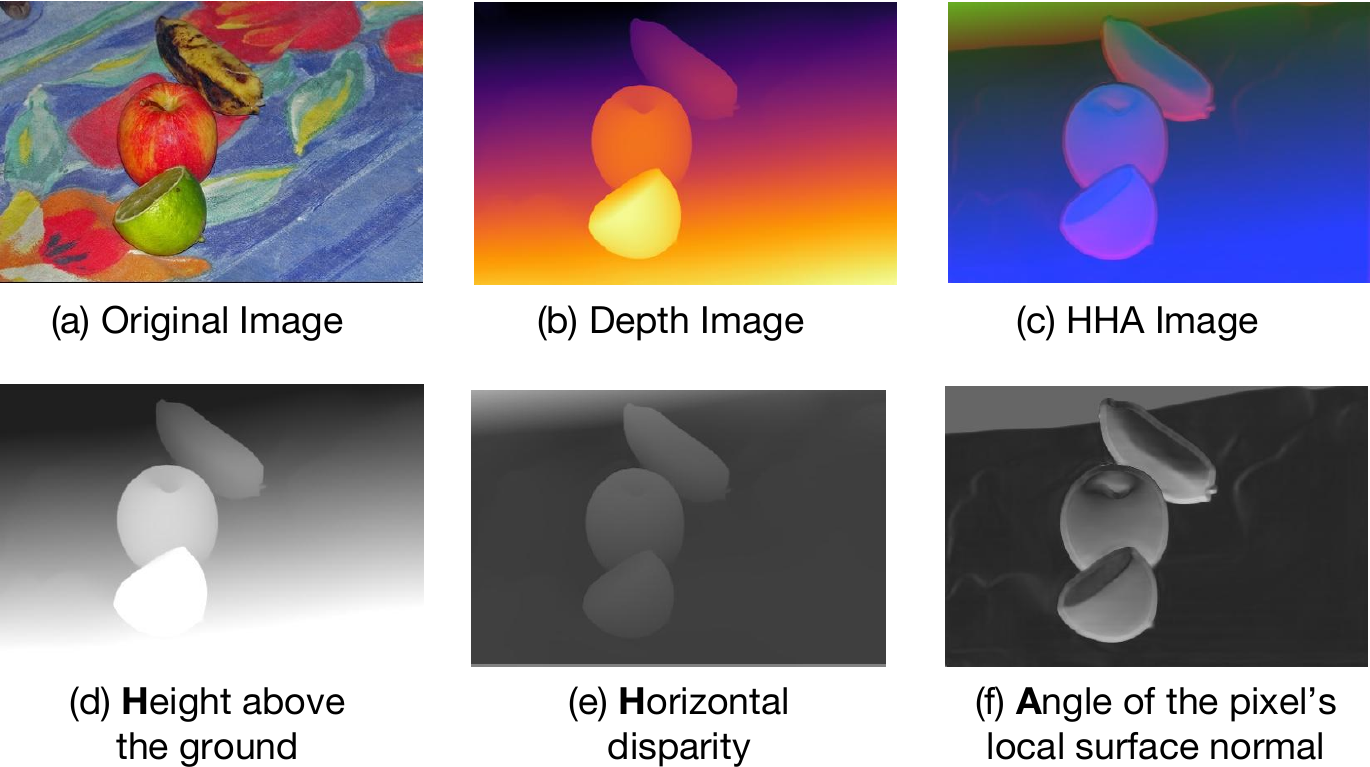}
    \caption{Example of images in our experiments.}
    \label{fig.hha}
\end{figure}
    
    

\subsection{Depth Guided HHA Representation Generation Module}
  As mentioned earlier, due to the complex information contained within an RGB image, one-stage scene graph generation models may struggle to effectively learn scene topology from it. To tackle this issue, we propose to utlize depth information.  In our work, We first adopt MiDas\cite{DBLP:journals/corr/abs-1907-01341} as our monocular depth estimator to predict depth maps, which subsequently serve as our guidance signal. Once the depth map is obtained, the most  straightforward idea is to directly extract deep feature from it to enhance the main scene graph generation network. However, the utility of monocularly predicted depth maps is hampered by two main factors. Firstly, these predictions inherently suffer from inaccuracies due to the ambiguity of depth perception from a single viewpoint, which can lead to erroneous estimations of the spatial relations between objects. Secondly, the extraction of meaningful features from depth maps is a nontrivial task due to their high-dimensional nature and the complex interdependencies between depth values across the image. 

  To this end, inspired by the methodology outlined in Gupta et al.\cite{gupta2014learning}, we propose to use the HHA format (Horizontal disparity, Height above ground, and the Angle the pixel's local surface normal makes with the inferred gravity direction). However, the generation of HHA images requires camera parameters. It is nearly impossible to obtain accurate camera parameters from existing scene graph generation datasets. To circumvent this issue, we introduce a simplified approach, drawing on common methods found in contemporary human pose estimation models \cite{li2021hybrik,zhang2021pymaf}. Specifically, we standardize the use of the image center as the camera origin and set a fixed focal length, despite the inherent limitations of this method. As a result, we can express our depth information, $\hat{I}$, as follows: $\hat{I} = (I_{\text{hori}}, I_{\text{height}}, I_{\text{angle}})$.
  
Fig.\ref{fig.hha} shows an illustration of the proposed the HHA presentation.  The HHA format provides a more intuitive understanding of the scene topology compared to raw depth maps, thereby offering a more effective guidance signal for scene graph generation.

\subsection{Depth Guided Semi-Teaching Network Learning Module }

For one-stage scene graph generation, we follow the definition provided by FCSGG \cite{liu2021fully}. This approach necessitates a single network to simultaneously perform object detection and relation prediction. We adopt the methodology of CenterNet \cite{duan2019centernet} to predict the center point and corresponding offset, $o$, in both  object detection branch and relation prediction branch.

To effectively leverage depth information, we propose a novel teacher-student model. This model enables the RGB-based network (student network) to learn depth information from the depth-based network (depth teacher). In order to circumvent the potential impact of missing information on model performance, we use depth information to supervise the intermediate offset, $o$, without imposing any additional constraints at the image feature level. This strategy is referred to as our semi-teacher module.

The depth teacher is developed using a deformation neural network. Here, the offset location serves as the teaching intermediate variable, denoted as $o_{i,j}^l$ for the offset at location $(i,j)$ of convolution kernel $l$. The depth teacher is initially trained with HHA (Horizontal disparity, Height above ground, Angle with gravity) representation as input. The HHA encoding effectively leverages depth information, enabling the depth teacher to more robustly understand the topological aspects of the scene. The output from this phase is the offset variable $o_{i,j}^l$ and can be represented mathematically as:

\begin{equation}
G_{depth}(HHA) = o_{i,j}^l
\end{equation}

Given that the HHA representation is also 3-channel like the original RGB image, the depth teacher network can share the exact same network architecture with the student network. This enhances consistency for the subsequent semi-teaching process.

The predicted offset is then regarded as the transferred knowledge, used to train the student network, which takes the original image as input, aiming to predict all parameters of a scene graph.

\begin{table*}
    \centering
    \scalebox{0.85}{
    \begin{tabular}{cccccccccccccc}
        \toprule
       \multirow{3}{*}{} & \textbf{Method} & \textbf{Backbone} & \multicolumn{3}{c}{\textbf{Prediction Classification}} & \multicolumn{3}{c}{\textbf{Scene Graph Classification}} & \multicolumn{3}{c}{\textbf{Scene Graph Detection}} & \textbf{img/sec}&\textbf{AP50} \\
        \cmidrule(lr){4-6} \cmidrule(lr){7-9} \cmidrule(lr){10-12}
        \multirow{8}{*}{\rotatebox{90}{\makecell{\textbf{Top-}\\\textbf{Down}}}}& & & \makecell[c]{\textbf{g-R20/}\\\textbf{ng-R20}}& \makecell[c]{\textbf{g-R50/}\\\textbf{ng-R50}}& \makecell[c]{\textbf{g-R100/}\\\textbf{ng-R100}}  & \makecell[c]{\textbf{g-R20/}\\\textbf{ng-R20}} & \makecell[c]{\textbf{g-R50/}\\\textbf{ng-R50}}&\makecell[c]{\textbf{g-R100/}\\\textbf{ng-R100}}  &\makecell[c]{\textbf{g-R50/}\\\textbf{ng-R50}}& \makecell[c]{\textbf{g-R50/}\\\textbf{ng-R50}} & \makecell[c]{\textbf{g-R100/}\\\textbf{ng-R100}}   \\
        \midrule
        & VC-Tree*\cite{yang2018graph} & VGG16\cite{simonyan2014very} & 60.1/- & 66.4/- & 68.1/- & 35.2/- & 38.1/- & 38.8/- & 22.0/- & 27.9/- & 31.3/- & 4.6&- \\
        & Seq2Seq*\cite{lu2021context} & VGG16\cite{simonyan2014very} & 60.3/- & 66.4/83.6 & 66.4/90.8 & 34.5/- & 38.3/46.9 & 39.0/68.5 & 22.1/- & 30.9/38.3 & 34.4/37.0 & -&- \\
        & BGNN*\cite{li2021bipartite} & RN101-FPN & -/- & 59.2/- & 61.3/- & -/- & 37.4/- & 38.5/- & -/- &31.0/- & 35.8/- & 1.6&- \\
        \midrule
        \multirow{8}{*}{\rotatebox{90}{\textbf{Bottom-Up}}} & Pix2Graph\cite{newell2017pixels} & Hg-104\cite{kim2020lightweight}& -/- & 82.0/- & 86.4/- & -/- & 35.7/- & 38.4/- & -/- & 15.5/- & 18.8/- & 0.18&- \\
        & FCSGG\cite{liu2021fully} & HRNet-48\cite{wang2020deep} & 27.6/32.2 & 34.9/46.3 & 38.5/56.6 & 12.3/13.5 & 15.5/19.3 & 17.2/23.6 & 11.0/12.4 & 15.1/18.2 & 18.1/23.0 & 14.2&25.0 \\
        & FCSGG\cite{liu2021fully} & HRNet-32\cite{wang2020deep} & 24.2/28.1 & 31.0/40.3 & 34.6/50.0 & 13.6/14.2 & 17.1/19.6 & 18.8/24.0 & 11.5/12.7 & 15.5/18.3 & 18.4/23.0 & 15.6 &21.6\\
        & FCSGG \cite{liu2021fully}& RN50-FPN×2 & 28.0/31.6 & 31.0/40.3 & 34.6/50.0 & 12.3/13.5 & 15.5/19.3 & 17.2/23.6 & 11.4/12.2 & 15.7/18.0 & 19.0/22.8 & 27.4&23.0\\
        & CoRF\cite{adaimi2023composite} & RN50 & 36.0/40.6 & 42.3/53.9 &44.1/62.4 & 13.6/13.7 & 14.8/18.3 & 14.9/20.6 & 11.6/13.9 & 14.5/17.6 & 15.9/19.9 & \textbf{33.3} &19.6\\
        & CoRF+T\cite{adaimi2023composite} & RN50 & 38.2/43.4 & 44.4/56.8 & 46.0/65.1 &15.9/16.0 & 17.2/21.3 & 17.4/23.6 & 13.3/15.9 & 17.5/20.2 & 18.1/22.6 & 20 &21.9\\
        & CoRF \cite{adaimi2023composite}& Swin-S\cite{liu2021swin} & 38.5/43.5 & 44.8/56.9 & 34.6/50.0 & 16.1/16.4 & 17.5/21.6 & 1.7/23.9 & 14.5/17.4 & 17.9/22.0 & 19.4/24.5 & 19.2 &23.8\\
        & CoRF+T \cite{adaimi2023composite}& Swin-S\cite{liu2021swin} & \textbf{39.3/44.9}& \textbf{45.4/58.1} & \textbf{47.1/66.3} & 17.3/17.7 & 18.7/23.4 &18.9/25.8 & 15.3/18.0& 18.6/22.9 & 20.0/25.4 & 17.2 &24.7\\
        \midrule
       \multirow{2}{*}{\rotatebox{90}{\textbf{Ours}} } & \textbf{STDG} & RN50 & 37.1/42.0 & 43.3/55.4 & 45.1/63.6 & 14.2/15.0 & 15.7/19.9 & 15.8/22.0 & 12.3/15.0 & 15.3/18.7 & 16.6/21.0 & \textbf{33.3}&20.7\\
        & \textbf{STDG} & Swin-S\cite{liu2021swin} & 39.1/44.2 & 45.3/57.6 & 46.9/65.7 & \textbf{18.2/18.7} & \textbf{19.7/24.4} & \textbf{19.8/26.8} & \textbf{15.7/18.6} & \textbf{19.1/23.4} & \textbf{20.5/26.0} & 20.8&\textbf{25.5}\\
        \bottomrule
    \end{tabular}
    }
    \caption{Main experimental results, computed on an A6000 GPU, with the speed reported as the average on the SGDet test dataset. AP50 indicates the accuracy of bounding box predictions with IOU $>$ 0.5. RN50 refers to Resnet50\cite{he2016deep}, and Swin-S represents Swin Transformer\cite{liu2021swin}. FPN signifies Feature Pyramid Net, and FPNx2 denotes the use of two FPNs for object detection and relationship prediction. Hg stands for Hourglass\cite{kim2020lightweight}. An asterisk (*) signifies results obtained using ground truth object detection results for training.}
    \label{Mainresults}
    \vspace{-0.25in}

\end{table*}

\subsection{Depth Guided Scene Graph Generation Module}
Once the depth teacher network is fully trained, we proceed to train the student network. The student network takes the original image as input and aims to output the complete set of parameters for a scene graph. During this training process, the offset label is provided by the depth teacher as a pseudo-label, while the ground truth directly supervises the rest of the parameters.

\noindent \textbf{Object Detection.} In our object detection module, our foundation is the structural framework of CenterNet \cite{duan2019centernet}. Each object is characterized by its location, classification, and size. The object locations and classifications are jointly represented using a single heatmap, denoted as $h\in R^{C\times W\times H}$, where $C$ signifies the number of object classes. The sizes of objects are represented as $s \in R^{2\times W\times H}$. To effectively incorporate depth information into the object detection process, we introduce two deformable convolutional layers positioned between the visual backbone and the prediction layer. The object detection loss, denoted as $Loss_{det}$, is formulated as the sum of the squared differences between the ground truth and the predicted values of the object heatmap ($h$) and object size ($s$):

\begin{equation}
    Loss_{det}=(h_{gt}-h_{pred})^2+(s_{gt}-s_{pred})^2
\end{equation}

\noindent \textbf{Relation Classification.} For the scene graph generation, we use the CoRF method as outlined in Adaimi et al.\cite{adaimi2023composite}. This method allows us to predict multiple relations at a single location simultaneously. We represent a relation, denoted as $r$, in the scene graph as $r_{i,j}^p = [\alpha, x_{obj}, y_{obj}, x_{subj}, y_{subj}, s_{subj}, s_{obj}]$, where $\alpha$ signifies the confidence level of relation $p$ at location $(i,j)$ for $subj$ and $obj$. The $x$ and $y$ correspond to the coordinates of the subject and object, while $s_{subj}$ and $s_{obj}$ denote one-ninth of the minimal width or height for the subject and object, respectively. The relation can be represented as $r\in R^{7\times W\times H}$, where $W$ and $H$ are the width and height of the scene. The loss function for relation prediction, denoted as $Loss_{rel}$, is formulated as follows:

\begin{equation}
    Loss_{rel}=(r_{gt}-r_{pred})^2
\end{equation}

\noindent \textbf{Semi-Supervised Learning.} The semi-supervision loss, denoted as $Loss_{semi}$, is calculated by comparing the predicted offsets of the student network and the depth teacher:

\begin{equation}
Loss_{semi}=\sum^W_{w=0}\sum^H_{h=0} (o_{i,j}^{w,h} -\hat{o}_{i,j}^{w,h})^2
\end{equation}

Here, $o$ and $\hat{o}$ represent the offsets generated by the student model and the depth teacher, respectively. Note, in both detection branch and relation prediction branch, there exist offset

\noindent \textbf{Overall Loss.} Finally, the overall loss for training the student network with an RGB image is calculated as follows:

\begin{equation}
 Loss=Loss_{det}+Loss_{rel}+Loss_{semi}*mode   
\end{equation}

Here, the $mode$ is set to 0 when training the depth model and 1 otherwise. This ensures that the semi-supervision loss is only incorporated when training the student model, reinforcing the learned depth information.

\section{Experiments}
    \subsection{Experiments Settings}
   Our experiments use the VG-100k dataset, a subset of VG-150K. This dataset comprises 108,000 images, 150 object categories, and 50 predicate categories. To validate our model, we utilize two backbone architectures: Resnet50 and Swin-Transformer. The representation of relations is based on the composite relation field method used for one-stage relation classification. During the training process, we set the backbone's learning rate to 5e-5 and the prediction part of the network to 5e-4. We apply gradient clipping at a rate of 5e-5. Both the deep network $G_{depth}$ and the network $G$ are trained for 60 epochs, with the last epoch results serving as our evaluation metrics. Evaluation is achieved through three aspects: Scene Graph Relationship Prediction, Scene Graph Predication Classification (PredCls), and Scene Graph Classification (SGCls). We utilize Recall@K as our primary metric, considering the VG-100k dataset's partially annotated nature. Additionally, we report non-graph constraints (ng-Recall@K) results and mean-Recall results to accommodate multiple possible relationships and the dataset's long-tail data.

\subsection{ Performance Analysis \& Comparison}

Our model's results are presented in three categories: recall performance, mean recall performance, and zero-shot recall performance, showcasing the superiority of our proposed semi-teaching depth-guided method.

The recall performance (Table.\ref{Mainresults}) illustrates our model's high inference speed, comparable to CoRF, and an approximate 1-unit increase in recall in our fastest mode. Our approach surpasses CoRF in terms of processing speed by 1.6 images per second and enhances SGDet's Recall50 by 1.2. Despite a slight performance dip in PreCls compared to the Transformer-enhanced CoRF model, substantial improvements are observed in SGCls and SGDet, alongside a computation speed increase of 3.6 images per second.

In mean recall performance (Table.\ref{meanrecall}), our model outperforms CoRF with the same backbone, presenting a maximum recall improvement of 1.4 in PredCLS. Despite a 1.1 decrease in ng-recall in PredCLS, other aspects did not drop by more than 0.3, and SGDet's recall even improved by 0.1 compared to CoRF+T.
\begin{table}
\centering

 \scalebox{0.9}{
\begin{tabular}{cccccccc}
\toprule
       Method& \multirow{2}{*}{Backbone} & \multicolumn{2}{c}{PredCLS} & \multicolumn{2}{c}{SGCLS} & \multicolumn{2}{c}{SGDet} \\
       \cmidrule(lr){3-4} \cmidrule(lr){5-6} \cmidrule(lr){7-8}
       &          & g       & ng      & g      & ng    & g     & ng    \\
\midrule
FCSGG  &  HRNet-48  & 5.5  & 9.7   & 2.5   & 4.4   & 2.4   & 3.6 \\
FCSGG  &  HRNet-32    & 5.2     & 9.5    & 2.9   & 6.3   & 2.6   & 4.7     \\
FCSGG  & RN50-FPNx2   & 5.7    & 11.3  & 2.9    & 6.0   & 2.7  & 4.9   \\
CoRF   & RN50  & 8.1    & 17.0  & 2.7& 5.4     & 2.7   & 5.8    \\
CoRF+T & RN50  & 9.5    & 20.0  & 3.4& 6.8   & 3.5    & 7.6 \\
CoRF   & Swin-S & 9.3     & 19.2  & 3.3 & 6.9    & 3.5   & 7.9   \\
CoRF+T & Swin-S & \textbf{10.1}   & \textbf{21.7}   &\textbf{ 3.9}   & \textbf{8.3}  & 3.9  & \textbf{9.2}   \\
\midrule
SDTG    & RN50  &   9.17&18.42&3.09&6.31&3.06&6.74 \\
SDTG    & Swin-S  & 9.8    & 20.6 & \textbf{3.9}  & 8.0   & \textbf{4.0}   & 9.0\\
\bottomrule


\end{tabular}
}

\caption{Mean Recall Performance. mR50 metric of different methods is shwon.}
\label{meanrecall}
\end{table}
For zero-shot recall performance, our model outperforms CoRF with ResNet50 as the backbone, improving ng recall50 in PredCLS by 1.0. Experiments with Swin Transformer enhanced recall by 0.3/0.6 in SGCLS, with a minor recall decrease of 0.2/0.1 in SGDet.
\begin{table}
\centering
 \scalebox{1}{
\begin{tabular}{cccccccc}
\toprule
\multirow{2}{*}{Method} & \multirow{2}{*}{Backbone} & \multicolumn{2}{c}{PredCLS} & \multicolumn{2}{c}{SGCLS} & \multicolumn{2}{c}{SGDet} \\
\cmidrule(lr){3-4} \cmidrule(lr){5-6} \cmidrule(lr){7-8}
       &          & g      & ng     & g     & ng     & g     & ng   \\
\midrule
FCSGG  & HRNet48         & 5.5          & 9.7          & 2.5         & 4.4         & 2.4         & 3.6         \\
FCSGG  & HRNet32         & 5.2          & 9.5          & 2.9         & 6.3         & 2.6         & 4.7         \\
FCSGG  &  RN50-FPN×2        & 5.7          & 11.3         & 2.9         & 6.0         & 2.7         & 4.9         \\
CoRF   & RN50     & 10.5         & 16.3         & 1.5         & 3.2         & 0.3         & 1.1         \\
CoRF+T & RN50     & 11.6         & 18.2         & 1.8         & 4.0         & 0.8         & 1.4         \\
CoRF   & Swin-S   & 11.1         & 18.0         & 1.9         & 3.5         & 1.1         & 2.2         \\
CoRF+T & Swin-S   & 11.3         & \textbf{18.8}         & 1.9         & 3.8         &\textbf{1.2}  & \textbf{2.6}         \\
\midrule
SDTG   & RN50     & 11.1         & 17.3         & 1.7         & 3.3         & 0.7         & 1.3         \\
SDTG   & Swin-S     & \textbf{11.7}         & 18.3         & \textbf{2.2}         & \textbf{4.4 }        & 1.0         & 2.5 \\
\bottomrule
\end{tabular}
}
\caption{Mean Recall Performance. zero-Recall50 of different methods is shown.}

\label{zero_shot}
\end{table}
Our model complexity analysis (Table.\ref{parameter}) reveals that our SDTG model has significantly fewer parameters compared to CoRF, making it more appealing for applications with lower GPU and CPU performance demands.
\begin{table}
\centering
 \scalebox{0.95}{
\begin{tabular}{cccccc}
\toprule
Method & Backbone & \makecell{Time per \\Image (ms)} &\makecell{ Images per\\ Second }& GMACS & \makecell{Params\\ (Millions)} \\
\midrule
CoRF   & RN50     & 30                   & 33.3               & 59.4 & 49.6          \\
CoRF+T & RN50     & 50                   & 20.0               & 50.2 & 40.7         \\
CoRF   & Swin-S   & 52                   & 21.3             & 68.5 & 69.3         \\
CoRF+T & Swin-S   & 59                   & 17.9              & 70.8 & 71.5          \\
\midrule
SDTG   & RN50     & 30                   & \textbf{33.3}               & 35.4 & \textbf{26.2}          \\
SDTG   & Swin-S   & 48                   & 20.8              & 62.6 & 63.5          \\
\bottomrule
\end{tabular}
}
\caption{Model Performance Metrics}
\label{parameter}

\end{table}

Overall, our proposed semi-teaching depth-guided method demonstrates superior performance in multiple aspects, thereby affirming its effectiveness in improving recall, mean recall, and zero-shot recall performances, and reducing model complexity.
   
    \subsection{Ablation Study}
    In this section, we investigate the effectiveness of each proposed module. We choose ResNet50 as the backbone. Results are shown as Table.\ref{ablation}. 

    \begin{table}
\centering
\begin{tabular}{cccc}
\toprule
\textbf{Method}                     & \textbf{PredCls}    & \textbf{SGCls}     & \textbf{SGDet}     \\
\midrule
Combine Training                    & 42.16/53.79         & 13.63/16.49        & 12.10/14.86        \\
w/o. Detection Guide             & 43.67/55.86         & 15.46/18.95        & 14.31/17.51        \\
w/o. Relation Guide              & 43.21/55.19         & 16.33/20.02        & 15.02/18.58        \\
R(HHA $\rightarrow$ depth)                    & 43.01/54.81         & 14.95/18.45        & 14.22/17.68        \\
w/. H                           & 42.78/54.73         & 14.80/18.15        & 13.90/17.06          \\
w/. HH                           & 42.72/54.82         & 15.20/18.62        & 14.43/17.80           \\
Fully teaching                 &36.34/47.3         &6.06/7.65         &    3.86/5.02 \\
\textbf{SDTG}                       & \textbf{43.28/55.35} & \textbf{15.69/19.86} & \textbf{15.25/18.71} \\
\bottomrule

\end{tabular}
\caption{Results of Ablation Study . \textit{Combine Training} means training the depth network and the rgb net work at the same time. \textit{w/o.} means that the experiments are conducted with out the guidance of detection or relation. \textit{$R(HHA \rightarrow depth)$ } means that the HHA features are replaced by the original depth information. \textit{w/. H} and \textit{w/. HH} means we use horizontal disparity only and use both horizontal disparity and height above the ground.}
\label{ablation}
\vspace{-0.2in}
\end{table}

    \textbf{Effectiveness  of Depth Guided Semi-Teaching Network Learning Module.} To investigate the effectiveness of depth guidance, we conduct an ablation study in which both modules were trained simultaneously while allowing the depth module to supervise the RGB scene graph generation model. The outcomes of this experiment are presented in the first row of Table.\ref{ablation}. The results indicate a substantial performance decline when the depth model is co-trained with the RGB model, thereby underscoring the crucial role of pretraining the depth extraction model in enhancing the overall effectiveness of the scene graph generation process. We also investigate the difference between fully-teach and semi-teach, it can be seen from  Table.\ref{ablation} that our semi-teach scheme achieves much better performance. 

    \textbf{Effectiveness of  Depth Guided Scene Graph Generation Module.}  To substantiate the effectiveness of depth guidance in both object detection and relation extraction, we conducted dedicated experiments: one with guidance training solely for object detection, and another exclusively for relation classification. As demonstrated in the second section of Table.\ref{ablation}, the absence of guidance in the object detection module enhances performance in PredCLS, underlining the model's proficiency in learning relation distributions. Conversely, without guidance in the relation classification module, the model exhibits improved performance in the SGCls task, signifying its strength in object detection. Optimal performance in SGDet tasks — a holistic metric for scene graph prediction that integrates both relation prediction and object detection — is only achieved when guidance is simultaneously employed in both object detection and relation classification. This underscores the importance of coordinated guidance in these two domains for superior scene graph prediction.

    \textbf{Effectiveness of Depth Guided HHA Representation Generation Module.} In order to verify the effectiveness of our approach to depth information processing using HHA data, we conducte an additional set of ablation studies. We experiment with raw depth information, horizontal disparity, and a combination of horizontal disparity with height above ground. As indicated in the fourth row of Table.\ref{ablation}, the SDTG model reaps tangible benefits from the enhancement of depth information using HHA features. Conversely, the utilization of solely horizontal disparity or height above ground information does not yield the superior performance associated with HHA data, underscoring the significant role HHA data plays in optimizing our model.


\section{Conclusion}
 In this work, we introduce the Depth-Guided One-Stage Scene Graph Generation (STDG) methodology, addressing existing challenges in scene graph generation for autonomous robotic systems. STDG's innovative architecture comprising three modules leverages depth information from generation to prediction without extra computational demand. Experimental results confirm its significant performance enhancement in one-stage scene graph generation.







\bibliographystyle{IEEEtran}
\bibliography{IEEEabrv,ref}

\end{document}